\documentclass[10pt,twocolumn,letterpaper]{article}

\usepackage{cvpr}
\usepackage{times}
\usepackage{epsfig}
\usepackage{graphicx}
\usepackage{amsmath}
\usepackage{amssymb}
\usepackage{algorithm}
\usepackage{algpseudocode}
\usepackage{float}
\usepackage{color}
\usepackage{colortbl}
\usepackage{breqn}
\usepackage{rotating}
\usepackage{times}
\usepackage{enumitem}
\usepackage{makecell}
\usepackage{lineno}
\usepackage{tabularx}
\usepackage{xcolor}
\usepackage{multirow}
\usepackage{graphicx}
\usepackage{hhline}
\usepackage{amsmath}
\usepackage{amssymb}
\usepackage{amsfonts}
\usepackage{empheq}
\usepackage{framed, color}
\usepackage{xcolor}
\usepackage{graphics}
\usepackage{graphicx}
\usepackage[]{graphicx} 
\usepackage{epsfig} 
\usepackage{color}
\usepackage{filecontents}
\usepackage{subfigure}
\usepackage{multirow}
\usepackage{filecontents}
\usepackage{verbatim} 
\usepackage{tabularx}
\usepackage{lineno}
\usepackage{setspace}
\usepackage{textcomp,booktabs}
\usepackage{caption}
\usepackage{stmaryrd}
\usepackage[mathscr]{eucal}
\usepackage{lineno}
\usepackage{subfigure}
\usepackage{adjustbox}
\usepackage[toc]{appendix}

\def\x{{\bf x}}
\def\y{{\bf y}}

\def\M{{\bf M}}

\def\0{{\bf 0}}
\def\1{{\bf 1}}

\def\RB{{\mathbb R}}

\definecolor{red}{rgb}{0.95,0.4,0.4}
\definecolor{blue}{rgb}{0.4,0.4,0.95}
\definecolor{darkblue}{rgb}{0,0,0.8}
\definecolor{darkred}{rgb}{0.8,0,0}
\definecolor{darkgreen}{rgb}{0,0.5,0}
\definecolor{grey}{rgb}{0.6,0.6,0.6}
\definecolor{col1}{RGB}{232, 161, 148}
\definecolor{col2}{RGB}{148, 187, 232}
\definecolor{lightgrey}{rgb}{0.85,0.85,0.85}

\usepackage[pagebackref=true,breaklinks=true,letterpaper=true,colorlinks,bookmarks=false]{hyperref}

\cvprfinalcopy 

\def\cvprPaperID{4343} 

\ifcvprfinal\fi
\begin{document}




\title{\textit{Domain Decluttering}: Simplifying Images to Mitigate \\ 
Synthetic-Real Domain Shift and Improve Depth Estimation}

\author{
Yunhan Zhao$^1$ \ \ Shu Kong$^2$ \ \ Daeyun Shin$^1$ \ \ Charless Fowlkes$^1$ \\
$^1$UC Irvine  \ \ \ \ \ \ \ \ \ \ $^2$Carnegie Mellon University \\
{\tt\small \{yunhaz5, daeyuns, fowlkes\}@ics.uci.edu } \ \ {\tt\small shuk@andrew.cmu.edu} \\ \\
}

\twocolumn[{%
\vspace{-1em}
\maketitle
\vspace{-3em}
\begin{center}
    \includegraphics[width=0.98\linewidth]{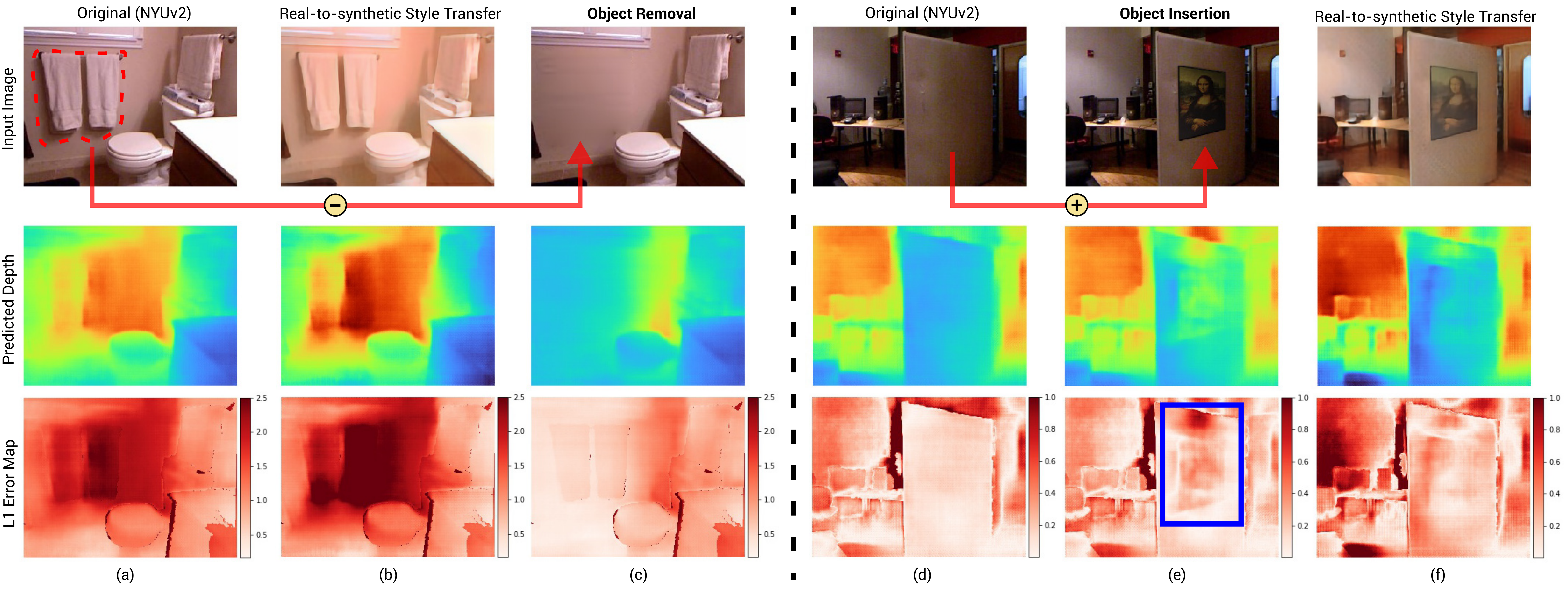}
\end{center}
\begin{center}
    \centering
    \vspace{-0.3in}
\captionof{figure}{\small
(a) The presence of novel objects and clutter can drastically degrade the
output of a well-trained depth predictor.
(b) Standard domain adaptation (\eg, a style translator trained with CycleGAN) only 
changes low-level image statistics and fails to solve the problem (even when trained with 
depth data from both synthetic and real domains), 
while
removing the clutter entirely (c) yields a remarkably better prediction.
Similarly, the insertion of a poster in (d,e) negatively affects the depth
estimate and low-level domain adaptation (f) only serves to hurt overall
performance.
}
\label{fig:splashy_figure}
\end{center}
}]

\begin{abstract}
Leveraging synthetically rendered data offers great potential
to improve monocular depth estimation and other geometric estimation tasks,
but closing the synthetic-real domain gap is a non-trivial and
important task.
While much recent work has focused on unsupervised domain adaptation,
we consider a more realistic scenario where a large amount of
synthetic training data is supplemented by a small set of real
images with ground-truth.
In this setting, we find that existing domain translation approaches
are difficult to train and offer little advantage over simple 
baselines that use a mix of real and synthetic data. A key failure
mode is that real-world images contain novel objects and clutter not
present in synthetic training. This high-level domain shift
isn't handled by existing image translation models.

Based on these observations,
we develop an attention module that learns to identify and remove difficult out-of-domain regions in real images in order to improve 
depth prediction for a model trained primarily on synthetic data.
We carry out extensive experiments to validate our 
attend-remove-complete approach (ARC) and find that it
significantly outperforms state-of-the-art domain adaptation methods 
for depth prediction. Visualizing the removed regions provides 
interpretable insights into the synthetic-real domain gap.
\end{abstract}

\vspace{-6mm}
\section{Introduction}

With a graphics rendering engine one can, in theory,
synthesize an unlimited number of scene images of 
interest and their corresponding ground-truth
annotations~\cite{zamir2018taskonomy, krahenbuhl2018free, zhang2016physically, song2016ssc}.
Such large-scale synthetic data increasingly serves as a source of training data
for high-capacity convolutional neural networks (CNN).
Leveraging synthetic data is particularly important for tasks such
as semantic segmentation that require fine-grained labels at each 
pixel and can be prohibitively expensive to manually annotate.
Even more challenging are pixel-level regression tasks where the 
output space is continuous. One such task, the focus of our paper,
is monocular depth estimation, where the only available ground-truth 
for real-world images comes from specialized sensors that typically 
provide noisy and incomplete estimates.

Due to the domain gap between synthetic and real-world imagery, 
it is non-trivial to leverage synthetic data.
Models naively trained over synthetic data 
often do not generalize well to the real-world images \cite{ganin2014unsupervised, long2015learning, tzeng2015simultaneous}.
Therefore domain adaptation problem has attracted increasing
attention from researchers aiming at closing the domain gap
through unsupervised generative models 
(\eg using GAN~\cite{goodfellow2014generative} or CycleGAN~\cite{zhu2017unpaired}).
These methods assume that domain adaptation can be largely resolved 
by learning a domain-invariant feature space or translating 
synthetic images into realistic-looking ones.
Both approaches rely on an adversarial discriminator to judge whether 
the features or translated images are similar across domains,
without specific consideration of the task in question.
For example,
CyCADA translates images between synthetic and real-world domains
with domain-wise cycle-constraints and adversarial learning~\cite{hoffman2017cycada}.
It shows successful domain adaptation for multiple vision tasks where
only the synthetic data have annotations while real ones do not.
T$^2$Net exploits adversarial learning to penalize the 
domain-aware difference between both images and features~\cite{zheng2018t2net},
demonstrating successful monocular depth learning where the synthetic data alone
provides the annotation for supervision.

Despite these successes, we observe two critical issues:

\noindent {\bf (1) Low-level vs. high-level domain adaptation}.
As noted in the literature~\cite{isola2017image, zhu2017toward},
unsupervised GAN models are limited in their ability to
translate images and typically only modify low-level factors, 
\eg, 
color and texture.
As a result,
current GAN-based domain translation methods 
are ill-equipped to deal with the fact that images from different domains contain 
high-level differences (\eg, novel objects present only in one domain),
that cannot be easily resolved.
Figure~\ref{fig:splashy_figure} highlights this difficulty.  High-level
domain shifts in the form of novel objects or clutter can drastically 
disrupt predictions of models trained on synthetic images. To combat this
lack of robustness, we argue that a better strategy may be to explicitly
identify and remove these unknowns rather than letting them corrupt 
model predictions.

\noindent {\bf (2)  Input vs. output domain adaptation}.
Unlike domain adaptation for image classification where appearances change
but the set of labels stays constant, in depth regression the domain shift is 
not just in the appearance statistics of the input (image) but also in 
the statistics of the output (scene geometry).
To understand how the statistics of geometry shifts between synthetic and 
real-world scenes, it is necessary that we have access to at least some 
real-world ground-truth. This precludes solutions that rely entirely on
unsupervised domain adaptation. However, we argue that a likely scenario is 
that one has available a small quantity of real-world ground-truth along 
with a large supply of synthetic training data.
As shown in our experiments, when we try to tailor existing unsupervised 
domain adaptation methods to this setup,
surprisingly we find that none of them perform satisfactorily and sometimes
even worse than simply training with small amounts of real data!

Motivated by these observations, we propose a principled approach that
improves depth prediction on real images using a somewhat unconventional
strategy of translating real images to make them more similar to 
the available bulk of synthetic training data.
Concretely,
we introduce an attention module that learns to 
detect problematic regions (\eg, novel objects or clutter) 
in real-world images.
Our attention module produces binary masks with the differentiable Gumbel-Max trick~\cite{gumbel2012statistics, jang2016categorical, veit2018convolutional, kong2019pixel},
and uses the binary mask to remove these regions from the input images.
We then develop an inpainting module that
learns to complete the erased regions with realistic 
fill-in.
Finally, a translation module adjusts the low-level factors such as 
color and texture to match the synthetic domain.

We name our translation model ARC, as it
attends, removes and completes 
the real-world image regions.
To train our ARC model,
we utilize a modular coordinate descent training pipeline where 
we carefully train each module individually and then fine-tune
as a whole to optimize depth prediction performance.
We find this approach is necessary since, as with other domain 
translation methods, the multiple losses involved 
compete against each other and do not necessarily contribute
to improve depth prediction.

To summarize our main contributions:
\begin{itemize}[leftmargin=*]
\vspace{-0.04in}
\setlength\parskip{-0.3em}
\item We study the problem of leveraging synthetic data along with a small amount of annotated real data 
for learning better depth prediction,
and reveal the limitations of current unsupervised domain
adaptation methods in this setting.
\item We propose a principled approach (ARC) that
learns identify, remove and complete 
``hard'' image regions in real-world images,
such that we can translate the real images 
to close the synthetic-real domain gap to improve 
monocular depth prediction.
\item We carry out extensive experiments to demonstrate the effectiveness of our ARC model,
which not only outperforms state-of-the-art methods,
but also offers good interpretability by explaining what 
to remove in the real images for better depth prediction. 
\end{itemize}

\section{Related Work}

\noindent \textbf{Learning from Synthetic Data}
is a promising direction in solving data scarcity,
as the render engine could in theory produce unlimited number 
of synthetic data and their perfect annotations 
used for training.
Many synthetic datasets have been released 
\cite{zhang2016physically, song2016ssc, gaidon2016virtual, levy2010sintel, butler2012naturalistic, dosovitskiy2015flownet},
for various pixel-level prediction tasks like
semantic segmentation, optical flow, 
and monocular depth prediction.
A large body of work uses synthetic data
to augment real-world datasets,
which are already large in scale,
to further improve the performance~\cite{li2018cgintrinsics, dosovitskiy2015flownet, varol2017learning}.
We consider a problem setting in which only a limited set of 
annotated real-world training data is available along with 
a large pool of synthetic data.

\noindent \textbf{Synthetic-Real Domain Adaptation}.
Models trained purely on
synthetic data often suffer limited generalization~\cite{pan2009survey}. 
Assuming there is no annotated real-world data during training,
one approach is to close synthetic-real domain gap
with the help of adversarial training.
These methods learn either a domain invariant feature space or 
an image-to-image translator that maps between images from synthetic and real-world domains.
For the former, 
\cite{long2013transfer} introduces Maximum Mean Discrepancy to learn domain invariant features;
\cite{tzeng2014deep} jointly minimizes MMD and classification error to further improve domain adaptation performance;
~\cite{tzeng2017adversarial, tsai2018learning} apply adversarial learning to 
aligning source and target domain features;
~\cite{sun2016deep} proposes to match the mean and variance of domain features. 
For the latter,
CyCADA learns to translate images from synthetic and real-world domains
with domain-wise cycle-constraints and adversarial learning~\cite{hoffman2017cycada}.
T$^2$Net exploits adversarial learning to penalize the 
domain-aware difference between both images and features~\cite{zheng2018t2net},
demonstrating successful monocular depth learning where the synthetic data alone
provide the annotation for supervision.

\noindent \textbf{Attention and Interpretability.}
Our model utilizes a learnable attention mechanism similar to those that have been widely adopted in the
community~\cite{sun2003object, vaswani2017attention, gehring2017convolutional},
improving not only the performance for the task in question~\cite{nguyen2018weakly, kong2019pixel}, 
but also improving interpretability and robustness from various perspectives~\cite{andreas2016neural, anderson2018bottom, veit2018convolutional, fang2019modularized}.
Specifically, we utilize the Gumbel-Max trick~\cite{gumbel2012statistics, jang2016categorical, veit2018convolutional, kong2019pixel}, to learn 
binary decision variables in a differentiable training framework.
This allows for efficient training while producing easily interpretable
results that indicate which regions of real images introduce errors 
that hinder the performance of models trained primarily on synthetic data.

\begin{figure*}[th]
\begin{center}
    \includegraphics[width=0.85\textwidth]{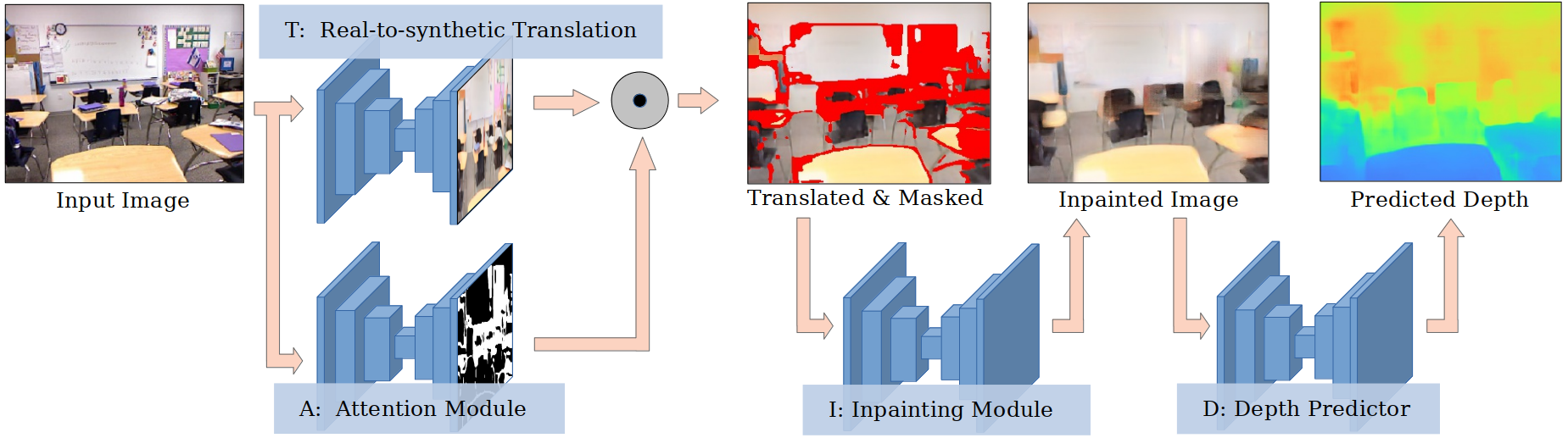}
\end{center}
\vspace{-6mm}
\caption{\small Flowchart of our whole ARC model in predicting the depth given a
real-world image.
The ARC framework performs real-to-synthetic translation
of an input image to account for low-level domain shift and 
simultaneously detects the ``hard'' out-of-domain regions using 
a trained attention module $\cal A$. These regions are 
removed by multiplicative gating with the binary mask from 
$\cal A$ and the masked regions inpainted by module $\cal I$. 
The translated result is fed to final depth predictor module 
$\cal D$ which is trained to estimate depth from a mix of 
synthetic and (translated) real data.
}
\vspace{-3mm}
\label{fig: system diagram}
\end{figure*}

\section{Attend, Remove, Complete (ARC)}
\label{sec:main}

Recent methods largely focus on how to leverage 
synthetic data (and their annotations) along with real-world images (where no annotations are available) to train a model that performs well on real images later~\cite{hoffman2017cycada, zheng2018t2net}.
We consider a more relaxed (and we believe realistic) scenario in
which there is a small amount of real-world ground-truth data available during training.
More formally, given a set of real-world labeled data 
$X^r=\{\x^r_i, \y^r_i\}_{i=1}^M$ and 
a large amount of synthetic data $X^s=\{\x^s_j, \y^s_j\}_{j=1}^N$,
where $M \ll N$, 
we would like to train a monocular depth predictor $\cal D$,
that accurately estimates per-pixel depth on real-world test images.
The challenges of this problem are two-fold.
First, due to the synthetic-real domain gap,
it is not clear when including synthetic training data improves the 
test-time performance of a depth predictor on real images.
Second, assuming the model does indeed benefit from synthetic training data,
it is an open question as to how to best leverage the knowledge of domain
difference between real and synthetic.

Our experiments positively answer the first question:
synthetic data can be indeed exploited for learning better depth models,
but in a non-trivial way as shown later through experiments.
Briefly, real-world images contain complex regions (\eg, rare objects),
which do not appear in the synthetic data. Such complex regions may negatively affect depth prediction by a model
trained over large amounts of synthetic, clean images.
Figure~\ref{fig: system diagram} demonstrates
the inference flowchart of {\bf ARC},
which learns to \emph{attend},  \emph{remove} and  
\emph{ complete} challenging
regions in real-world test images in order to better 
match the low- and high-level domain statistics of 
synthetic training data.
In this section, we elaborate each component module,
and finally present the training pipeline.

\subsection{Attention Module $\cal A$}

How might we automatically discover the existence and appearance of ``hard regions'' that negatively affect
depth learning and prediction? Such regions are not just those which are rare in the real images,
but also include those which are common in real images but absent from our pool of synthetic training data.
Finding such ``hard regions'' thus relies on both the depth predictor itself and synthetic data distribution.
To discover this complex dependence, we utilize an attention module $\cal A$
that learns to automatically detect such ``hard regions'' from the real-world input images.
Given a real image $\x^r \in \RB^{H\times W\times 3}$ as input, the attention module produces a binary 
mask $\M \in  \RB^{H\times W}$ used for erasing the ``hard regions'' using simple Hadamard product 
$\M\odot \x^r$ to produce the resulting masked image.

One challenge is that producing a binary mask typically involves a hard-thresholding operation
which is non-differentiable and prevents from end-to-end training using backpropagation. To solve this, we turn to the Gumbel-Max trick~\cite{gumbel2012statistics}
that produces quasi binary masks using a continuous
relaxation~\cite{jang2016categorical, maddison2016concrete}.

We briefly summarize the ``Gumbel max trick''~\cite{jang2016categorical, maddison2016concrete, veit2018convolutional}. A random variable $g$ follows a Gumbel distribution 
if $g = -\log( -\log (u))$, where $u$ follows a uniform distribution $U(0, 1)$. 
Let $m$ be a discrete binary random variable\footnote{A binary variable $m=0$ indicates the current pixel will be removed.} with probability $P(m = 1) \propto \alpha$, and let $g$ be a Gumbel random variable. 
Then, a sample of $m$ can be obtained by sampling $g$ from the Gumbel distribution and computing:
\begin{equation}
    {m} = \text{sigmoid}((\log(\alpha) + g)/\tau),
\end{equation}
where the temperature $\tau \rightarrow 0$ drives the ${m}$ to take on binary values and
approximates the non-differentiable argmax operation.
We use this operation to generate a binary mask of size $\M \in \RB^{H\times W}$.

To control the sparsity of the output mask $\M$, we penalize the empirical 
sparsity of the mask $\xi = \frac{1}{H*W} \sum_{i, j}^{H, W} \M_{i,j}$ 
using a KL divergence loss term~\cite{kong2019pixel}:
\begin{equation}
    \ell_{KL} = \rho \log(\frac{\rho}{\xi}) + (1 - \rho) \log(\frac{1 - \rho}{1 - \xi}).
\end{equation}
where hyperparameter $\rho$ controls the sparsity level (portion of pixels to keep).
We apply the above loss term $\ell_{KL}$ in training our whole system,
forcing the attention module $\cal A$ to identify the hard regions in an ``intelligent''
manner to target a given level of sparsity while still remaining the fidelity of 
depth predictions on the translated image.
We find that training in conjunction with the other modules results in attention masks
that tend to remove regions instead of isolated pixels (see Fig.~\ref{fig:visual_results})

\subsection{Inpainting Module $\cal I$} 
The previous attention module $\cal A$
removes hard regions in $\x^r$\footnote{Here, we present the inpainting module as a standalone piece. The final pipeline is shown in Fig.~\ref{fig: system diagram}} with sparse, binary mask $\M$,
inducing holes in the image with the operation $\M\odot \x^r$.
To avoid disrupting depth prediction we would like to fill in some 
reasonable values (without changing unmasked pixels).
To this end,
we adopt an inpainting module $\cal I$
that learns to fill in holes by leveraging knowledge from synthetic data distribution as well as the depth prediction loss.
Mathematically we have:
\begin{equation}
    \Tilde{\x} = (1 - \M) \odot {\cal I}(\M \odot \x^r) + \M \odot \x^r.
\end{equation}

To train the inpainting module $\cal I$,
we pretrain with a self-supervised method by learning
to reconstruct randomly removed regions using the reconstruction loss $\ell_{rec}^{rgb}$:
\begin{equation}
\ell_{rec}^{rgb} = 
        \mathbb{E}_{x^r \sim X^r}[||{\cal I}(\M \odot \x^r) 
        - \x^r||_1] 
\end{equation}
As demonstrated in~\cite{shetty2018adversarial}, $\ell_{rec}^{rgb}$ encourages the model to learn remove objects instead of reconstructing the original images since removed regions are random.
Additionally, 
we use two perceptual losses~\cite{zhang2018unreasonable}. 
The first penalizes feature reconstruction:
\begin{equation}
    \ell^f_{rec} = \sum\limits_{k=1}^K \mathbb{E}_{\x^r \sim X^r}
    [|| \phi_k({\cal I}(\M \odot \x^r)) - \phi_k(\x^r)||_1],
\end{equation}
where $\phi_k(\cdot)$ is the output feature 
at the $k^{th}$ layer of a VGG16 pretrained model~\cite{simonyan2014very}.
The second perceptual loss is a style reconstruction loss that penalizes the differences in colors, textures, and common patterns.
\begin{equation}
	\ell_{style} = \sum\limits_{k=1}^K \mathbb{E}_{x^r \sim X^r} [|| \sigma^{\phi}_k({\cal I}(\M \odot \x^r)) - \sigma^{\phi}_k(\x^r)||_1],
\end{equation} 
where function $\sigma^{\phi}_k(\cdot)$ returns a Gram matrix.
For the feature 
$\phi_k(\x)$ of size $C_k \times H_k \times W_k$, 
the corresponding Gram matrix $	\sigma^{\phi}_k(\x^r) \in \RB^{C_k \times C_k}$ 
is computed as:
\begin{equation}
	\sigma^{\phi}_k(\x^r) = \frac{1}{C_k H_k W_k} R(\phi_k(\x^r)) \cdot R(\phi_k(\x^r))^T,
\end{equation}
where $R(\cdot)$ reshapes the feature $\phi_k(\x)$ into $C_k \times H_k \ W_k$. 

Lastly, 
we incorporate an adversarial loss $\ell_{adv}$
to force the inpainting module $\cal I$ to fill in 
pixel values that follow the {\em synthetic} data distribution:
\begin{equation}
    \label{adversarial}
    \ell_{adv} = \mathbb{E}_{x^r \sim X^r}[\log(D(\Tilde{\x}))] + \mathbb{E}_{\x^s \sim X^s}[\log(1 - D(\x^s)],
\end{equation}
where $D$ is a discriminator with learnable weights that is trained on the fly.
To summarize,
we use the following loss function to train our inpainting module $\cal I$:
\begin{equation}
    \ell_{inp} = \ell_{rec}^{rgb} 
    + \lambda_f \cdot \ell^f_{rec} 
    + \lambda_{style} \cdot \ell_{style} 
    + \lambda_{adv} \cdot \ell_{adv},
\end{equation}
where we set weight parameters as $\lambda_f = 1.0$, $\lambda_{style}=1.0$, and $\lambda_{adv}=0.01$ in our paper.

\subsection{Style Translator Module $\cal T$} 

The style translator module $\cal T$ is the final piece 
to translate the real images into the synthetic data domain.
As the style translator adapts low-level feature 
(\eg, color and texture) we apply it prior to inpainting.
Following the literature,
we train the style translator in a standard CycleGAN~\cite{zhu2017unpaired} pipeline,
by minimizing the following loss:
\begin{equation}
\begin{split}
   \ell_{cycle} &= \mathbb{E}_{x^r \sim X^r}[|| G_{s2r}(G_{r2s}(x^r)) - x^r  ||_1] \\
    &+ \mathbb{E}_{x^s \sim X^s}[|| G_{r2s}(G_{s2r}(x^s)) - x^s  ||_1], 
\end{split}
\end{equation}
where $\cal T$ = $G_{r2s}$ is the translator from direction real to synthetic domain; 
while $G_{s2r}$ is the other way around. 
Note that we need two adversarial losses $\ell^r_{adv}$ and $\ell^s_{adv}$ in the form of Eqn.(\ref{adversarial}) along with the cycle constraint loss $\ell_{cycle}$. 
We further exploit the identity mapping constraint to encourage translators to preserve the geometric content and color composition between original and translated images:
\begin{equation}
    \begin{split}
    \ell_{id} &= \mathbb{E}_{x^r \sim X^r}[|| G_{s2r}(x^r) - x^r  ||_1] \\
    &+ \mathbb{E}_{x^s \sim X^s}[|| G_{r2s}(x^s) - x^s  ||_1].
    \end{split}
\end{equation}
To summarize,
the overall objective function for training the style translator $\cal T$ is:
\begin{equation}
    \ell_{trans} = \lambda_{cycle} \cdot \ell_{cycle} + \lambda_{id} \cdot \ell_{id} +  (\ell^r_{adv} + \ell^s_{adv}),
\end{equation}
where we set the weights $\lambda_{cycle}=10.0$, $\lambda_{id}=5.0$.

\subsection{Depth Predictor $\cal D$}

We train our depth predictor $\cal D$ over the combined set of translated real 
training images $\Tilde{\x}^r$ and synthetic images $\x^s$ using a 
simple $L_1$ norm based loss:
\begin{equation}
\begin{split}
\ell_d = &\mathbb{E}_{(\x^r, \y^r) \sim X^r}
||{\cal D}(\tilde{\x}^r) - y^r||_1  \\
&+ \mathbb{E}_{(\x^s, \y^s) \sim X^s} ||{\cal D}(\x^s) - \y^s||_1.
\end{split}
\end{equation}

\subsection{Training by Modular Coordinate Descent}

In principle, one might combine all the loss terms to train the ARC
modules jointly.  However, we found such practice difficult
due to several reasons:
bad local minima,
mode collapse within the whole system,
large memory consumption, etc.
Instead,
we present our proposed training pipeline
that trains each module individually,
followed by a fine-tuning step over the whole system.
We note such a modular coordinate descent training protocol 
has been exploited in prior work,
such as block coordinate descent methods~\cite{mairal2009online, aharon2006k},
layer pretraining in deep models~\cite{hinton2006reducing, simonyan2014very},
stage-wise training of big complex systems~\cite{barshan2015stage, chen2018universal} 
and those with 
modular design~\cite{andreas2016neural, fang2019modularized}.

Concretely,
we train the depth predictor module $\cal D$ by feeding the 
original images from either the synthetic set,
or the real set or the mix of the two as a pretraining stage.
For the synthetic-real style translator module $\cal T$,
we first train it with CycleGAN.
Then we insert the attention module $\cal A$ and the depth predictor
module $\cal D$ into this CycleGAN,
but fixing $\cal D$ and $\cal T$,
and train the attention module $\cal A$ only.
Note that after training the attention module $\cal A$,
we fix it without updating it any more and switch
the Gumbel transform to output real binary maps,
on the assumption that it has already
learned what to attend and remove with the help of depth loss 
and synthetic distribution.
We train the inpainting module $\cal I$ over 
translated real-world images and synthetic images.

The above procedure yields good initialization for all the modules,
after which we may keep optimizing them one by one while fixing 
the others.
In practice,
simply fine-tune the whole model 
(still fixing $\cal A$)
with the depth loss term only, by removing all the adversarial losses.
To do this,
we alternate between minimizing the following two losses:
\begin{equation}
    \label{update I, F, T}
    \begin{split}
        \ell^1_d &= \mathbb{E}_{(\x^r, \y^r) \sim X^r} ||{\cal D}({\cal I}({\cal T}(\x^r)\odot{\cal A}(\x_r)))) - \y^r||_1 \\
        &+ \mathbb{E}_{(\x^s, \y^s) \sim X^s} ||{\cal D}(\x^s) - \y^s||_1,
    \end{split}
\end{equation}
\begin{equation}
    \label{update F}
    \ell^2_d = \mathbb{E}_{(\x^r, \y^r) \sim X^r} ||{\cal D}({\cal I}({\cal T}(\x^r)\odot {\cal A}(\x_r)))) - \y^r||_1.
\end{equation}
We find in practice that 
such fine-tuning better exploits synthetic data to avoid overfitting
on the translated real images,
and also avoids catastrophic 
forgetting~\cite{french1999catastrophic, kirkpatrick2017overcoming} on the real images 
in face of overwhelmingly large amounts of synthetic data.


\section{Experiments}
\label{sec:exp}

We carry out extensive experiments
to validate our ARC model in leveraging synthetic data
for depth prediction.
We provide systematic ablation study to 
understand the contribution of each module
and the sparsity of the attention module $\cal A$.
We further visualize the intermediate results
produced by ARC modules,
along with failure cases,
to better understand the whole ARC model
and the high-level domain gap.

{
\setlength{\tabcolsep}{0.5em} 
\begin{table}[t]
\caption{A list of metrics used for evaluation in experiments,
with their calculations,
denoting by $y$ and $y^{*}$ the predicted and ground-truth
depth in the validation set.}
\vspace{-3mm}
\centering
\scriptsize
{
\begin{tabular}{| l | l l | }
\hline
Abs Relative diff. (Rel)   & &  $\frac1{|T|}\sum_{y\in T}|y - y^*| / y^*$ \\
Squared Relative diff. (Sq-Rel) & & $\frac1{|T|}\sum_{y\in T}||y - y^*||^2 / y^*$ \\
RMS & & $\sqrt{\frac1{|T|}\sum_{y\in T}||y_i - y_i^*||^2}$ \\
RMS-log & &   $\sqrt{\frac1{|T|}\sum_{y\in T}||\log y_i - \log y_i^*||^2}$ \\
Threshold  $\delta^i$, \ \ $i$$\in$$\{1,2,3\}$ &  &  \% of $y_{i}$ s.t. $\max(\frac{y_i}{y_i^*},\frac{y_i^*}{y_i})$$<$$1.25^i$ \\
\hline
\end{tabular}
}
\label{tab:metrics}
\vspace{-4mm}
\end{table}
}

\subsection{Implementation Details}

\noindent \textbf{Network Architecture}.
Every single module in our ARC framework is implemented
by a simple encoder-decoder architecture as used in ~\cite{zhu2017unpaired}, 
which also defines our discriminator's architecture.
We modify the decoder to output a single channel to train
our attention module $\cal A$.
As for the depth prediction module,
we further add skip connections that help output
high-resolution depth estimate~\cite{zheng2018t2net}.

\noindent \textbf{Training}.
We first train each module individually for 50 epochs using the Adam optimizer~\cite{kingma2014adam}, 
with initial learning rate 5$e$-5 (1$e$-4 for discriminator if adopted) and coefficients 0.9 and 0.999
for computing running averages of gradient and its square.
Then we fine-tune the whole ARC model with the proposed modular coordinate descent scheme with the same learning parameters.

\noindent\textbf{Datasets}.
We evaluate on indoor scene and outdoor scene datasets.
For indoor scene depth prediction,
we use the real-world NYUv2~\cite{silberman2012indoor} and synthetic Physically-based Rendering (PBRS)~\cite{zhang2016physically} datasets.
NYUv2 contains video frames captured using Microsoft Kinect,
with 1,449 test frames and a large set of video (training) frames.
From the video frames, 
we randomly sample 500 as our small amount of labeled real data (no overlap with the official testing set).
PBRS  contains large-scale synthetic images generated using the Mitsuba renderer and SUNCG CAD models~\cite{song2016ssc}.
We randomly sample 5,000 synthetic images for training.
For outdoor scene depth prediction,
we turn to the Kitti~\cite{geiger2013vision} and virtual Kitti (vKitti)~\cite{gaidon2016virtual} datasets.
In Kitti, we use the Eigen testing set to evaluate~\cite{zhou2017unsupervised, godard2019digging} and the first 1,000 frames as the small amount of real-world labeled data for training~\cite{eigen2014depth}.
With vKitti,
we use the split $\{$clone, 15-deg-left, 15-deg-right, 30-deg-left, 30-deg-right$\}$ to form 
our synthetic training set consisting of 10,630 frames. 
Consistent with previous work, 
we clip the maximum depth in vKitti to 80.0m for training,
and report performance on Kitti by capping at 80.0m for a fair comparison.

\noindent\textbf{Comparisons and Baselines}. We compare four 
classes of models. 
Firstly we have three baselines that train a single depth 
predictor on only synthetic data, only real data, or the 
combined set.
Secondly we train state-of-the-art domain adaptation
methods (T$^2$Net~\cite{zheng2018t2net}, 
CrDoCo~\cite{chen2019crdoco} and GASDA~\cite{zhao2019geometry})
with their released code. 
During training, we also modify them
to use the small amount of annotated real-world data
in addition to the large-scale synthetic data.
We note that these methods originally perform unsupervised
domain adaptation,
but they perform much worse than our modified baselines which leverage some real-world training data.
This supports our suspicion that multiple losses involved 
in these methods (\eg, adversarial loss terms)
do not necessarily contribute to reducing the depth loss.
Thirdly,
we have our ARC model and ablated variants to evaluate
how each module helps improve depth learning.
The fourth group includes a few top-performing fully-supervised
methods which were trained specifically for the dataset
over annotated real images only,
but at a much larger scale
For example,
DORN~\cite{fu2018deep} trains over more than 120K/20K frames
for NYUv2/Kitti, respectively.
This is 200/40 times larger than the 
labeled real images for training our ARC model.

\noindent\textbf{Evaluation metrics}.
for depth prediction are standard and widely adopted in literature,
as summarized  in Table~\ref{tab:metrics},

\subsection{Indoor Scene Depth with NYUv2 \& PBRS}

{
\setlength{\tabcolsep}{0.08em} 
\begin{table}[t]
\centering
\tiny
\caption{\small \textbf{Quantitative comparison} over 
NYUv2 testing set~\cite{silberman2012indoor}.
We train the state-of-the-art domain adaptation methods with the small amount
of annotated real data in addition to the large-scale synthetic data.
We design three baselines that only train  a single depth predictor directly over synthetic or real images.
Besides report \emph{full} ARC model, 
we ablate each module or their combinations.
We set $\rho$=0.85 in the attention module $\cal A$ if any,
with more ablation study in Fig.~\ref{fig: rho curve nyu}.
Finally, as reference,
we also list a few top-performing methods that 
have been trained over several orders more annotated real-world frames.
}
\vspace{-3mm}
\resizebox{0.48\textwidth}{!}{
\begin{tabular}{l| c c c c | c c c}
\hline
\multirow{2}{*}{Model/metric} &  \multicolumn{4}{c|}{\cellcolor{col1} \texttt{$\downarrow$ lower is better}} &  \multicolumn{3}{c}{\cellcolor[rgb]{0.0,0.8,1.0} \texttt{$\uparrow$ better}} \\
& \cellcolor{col1} { Rel} & \cellcolor{col1} { Sq-Rel} & \cellcolor{col1} { RMS} & \cellcolor{col1} { RMS-log}
& \cellcolor[rgb]{0.0,0.8,1.0}{$\delta^1$} 
& \cellcolor[rgb]{0.0,0.8,1.0}{$\delta^2$}  
& \cellcolor[rgb]{0.0,0.8,1.0}{$\delta^3$} \\
\hline
\multicolumn{8}{c}{\cellcolor{lightgrey}
State-of-the-art domain adaptation methods (\emph{w/ real labeled data})} \\
\hline
T$^2$Net~\cite{zheng2018t2net}  & {0.202} & {0.192} & {0.723} & {0.254} & {0.696} & {0.911} & {0.975} \\
CrDoCo~\cite{chen2019crdoco} & {0.222} & {0.213} & {0.798} & {0.271} & {0.667} & {0.903} & {0.974} \\
GASDA~\cite{zhao2019geometry} & {0.219} & {0.220} & {0.801} & {0.269} & {0.661} & {0.902} & {0.974} \\
\hline
\multicolumn{8}{c}{\cellcolor{lightgrey}
Our (baseline) models.} \\
\hline
syn only & 0.299 & 0.408 & 1.077 & 0.371 & 0.508 & 0.798 & 0.925 \\
real only & 0.222 & 0.240 & 0.810 & 0.284 & 0.640 & 0.885 & 0.967 \\
mix training & 0.200 & 0.194 & 0.722 & 0.257 & 0.698 & 0.911 & 0.975 \\
\hline
ARC: $\cal T$ & {0.226} & {0.218} & {0.805} & {0.275} & {0.636} & {0.892} & {0.974} \\
ARC:$\cal A$ & {0.204} & {0.208} & {0.762} & {0.268} & {0.681} & {0.901} & {0.971} \\
ARC: $\cal A$\&$\cal T$ & {0.189} & {0.181} & {0.726} & {0.255} & {0.702} & {0.913} & {0.976} \\
ARC: $\cal A$\&$\cal I$ & {0.195} & {0.191} & {0.731} & {0.259} & {0.698} & {0.909} & {0.974} \\
ARC: \emph{full} & {\bf 0.186} & {\bf 0.175} & {\bf 0.710} & {\bf 0.250} & {\bf 0.712} & {\bf 0.917} & {\bf 0.977} \\
\hline
\multicolumn{8}{c}{\cellcolor{lightgrey}
Training over large-scale NYUv2 video sequences ($>$120K)} \\
\hline
DORN~\cite{fu2018deep} & 0.115 & -- & 0.509 & 0.051 & 0.828 & 0.965 & 0.992 \\
Laina~\cite{laina2016deeper} & 0.127 & -- & 0.573 & 0.055 & 0.811 & 0.953 & 0.988 \\
Eigen~\cite{eigen2015predicting} & {0.158} & {0.121} & {0.641} & 0.214 & 0.769 & 0.950 &  0.988 \\
\hline
\end{tabular}
}
\label{tab:indoor}
\vspace{-4mm}
\end{table}
}

Table~\ref{tab:indoor} lists detailed comparison for indoor scene depth prediction.
We observe that ARC outperforms other unsupervised domain adaptation methods by a substantial margin.
This demonstrates two aspects.
First, these domain adaptation methods have adversarial losses that force translation between 
domains to be more realistic,
but there is no guarantee that ``more realistic'' is beneficial for depth learning.
Second, removing ``hard'' regions in real images makes the real-to-synthetic translation easier and 
more effective for leveraging synthetic data in terms of depth learning.
The second point will be further verified through qualitative results.
We also provide an ablation study
adding in the attention module $\mathcal{A}$ 
leads to better performance than merely adding the synthetic-real style translator $\mathcal{T}$.
This shows the improvement brought by $\cal A$.
However, combining $\cal A$ with either $\cal T$ or $\cal I$ 
improves further,
while  $\cal A$ \& $\cal T$ is better as removing the hard real regions
more closely matches the synthetic training data.

\subsection{Outdoor Scene Depth with Kitti \& vKitti}

We train the same set of domain adaptation methods and baselines on
the outdoor data,
and report detailed comparisons in Table~\ref{tab:outdoor}.
We observe similar trends as reported in the indoor scenario in Table~\ref{tab:indoor}.
Specifically,
$\cal A$ is shown to be effective in terms of better performance prediction;
while combined with other modules (\eg, $\cal T$ and $\cal I$)
it achieves even better performance.
By including all the modules, our ARC model (the \emph{full} version)
outperforms by a clear margin the other domain adaptation methods and the baselines.
However,
the performance gain here is not as remarkable as that in the indoor scenario.
We conjecture this is due to several reasons:
1) depth annotation by LiDAR are very sparse while vKitti have annotations everywhere;
2) the Kitti and vKitti images are far less diverse than indoor scenes (e.g., similar
perspective structure with vanishing point around the image center).

{
\setlength{\tabcolsep}{0.08em} 
\begin{table}[t]
\centering
\tiny
\caption{\small  \textbf{Quantitative comparison} over 
Kitti testing set~\cite{geiger2013vision}.
The methods we compare are the same as described in Table~\ref{tab:indoor},
including three baselines, our ARC model and ablation studies,
the state-of-the-art domain adaptation methods trained on both synthetic and real-world annotated data,
as well as some top-performing methods on this dataset,
which have been trained over three orders more annotated real-world frames from
kitti videos.
}
\vspace{-3mm}
\resizebox{0.48\textwidth}{!}{
\begin{tabular}{l|c c c c | c c c}
\hline
\multirow{2}{*}{Model/metric} 
&  \multicolumn{4}{c|}{\cellcolor{col1}\texttt{$\downarrow$ lower is better}} 
&  \multicolumn{3}{c}{\cellcolor[rgb]{0.0,0.8,1.0}\texttt{$\uparrow$ better}} \\
& \cellcolor{col1} { Rel} & \cellcolor{col1} { Sq-Rel} 
& \cellcolor{col1} { RMS} & \cellcolor{col1} { RMS-log}
& \cellcolor[rgb]{0.0,0.8,1.0} { $\delta^1$} 
& \cellcolor[rgb]{0.0,0.8,1.0} { $\delta^2$} 
& \cellcolor[rgb]{0.0,0.8,1.0} { $\delta^3$} \\
\hline
\multicolumn{8}{c}{\cellcolor{lightgrey}
State-of-the-art domain adaptation methods (\emph{w/ real labeled data})} \\
\hline
T$^2$Net~\cite{zheng2018t2net} & {0.151} & {0.993} & {4.693} & {0.253} & {0.791} & {0.914} & {0.966} \\
CrDoCo~\cite{chen2019crdoco} & {0.275} & {2.083} & {5.908} & {0.347} & {0.635} & {0.839} & {0.930} \\
GASDA~\cite{zhao2019geometry} & {0.253} & {1.802} & {5.337} & {0.339} & {0.647} & {0.852} & {0.951} \\
\hline
\multicolumn{8}{c}{\cellcolor{lightgrey}
Our (baseline) models.} \\
\hline
syn only & 0.291 & 3.264 & 7.556 & 0.436 & 0.525 & 0.760 & 0.881 \\
real only & 0.155 & 1.050 & 4.685 & 0.250 & 0.798 & 0.922 & 0.968 \\
mix training & 0.152 & 0.988 & 4.751 & 0.257 & 0.784 & 0.918 & 0.966 \\
\hline
ARC: $\cal T$ & {0.156} & {1.018} & {5.130} &
{0.279} & {0.757} & {0.903} & {0.959} \\
ARC: $\cal A$ & {0.154} & {0.998} & {5.141} & {0.278} & {0.761} & {0.908} & {0.962} \\
ARC: $\cal A$\&$\cal T$ & {0.147} & {0.947} & {4.864} & {0.259} & {0.784} & {0.916} & {0.966} \\
ARC: $\cal A$\&$\cal I$ & {0.152} & {0.995} & {5.054} & {0.271} & {0.766} & {0.908} & {0.962} \\
ARC: \emph{full}  & {\bf 0.143} & {\bf 0.927} & {\bf 4.679} & {\bf 0.246}  & {\bf 0.798} & {\bf 0.922} & {\bf 0.968} \\
\hline
\multicolumn{8}{c}{\cellcolor{lightgrey}
Training over large-scale kitti video frames ($>$20K)} \\
\hline
DORN~\cite{fu2018deep} & 0.071 & 0.268 & 2.271 & 0.116 & 0.936 & 0.985 & 0.995 \\
DVSO \cite{yang2018deep} & 0.097 & 0.734 & 4.442 & 0.187 & 0.888 & 0.958 & 0.980\\
Guo \cite{guo2018learning} &  0.096 & 0.641  & {4.095}  & {0.168}  & {0.892}  & {0.967} &0.986  \\
\hline
\end{tabular}
}
\label{tab:outdoor}
\vspace{-1mm}
\end{table}
}

\subsection{Ablation Study and Qualitative Visualization}


\begin{figure}[t]
\includegraphics[width=0.999\linewidth]{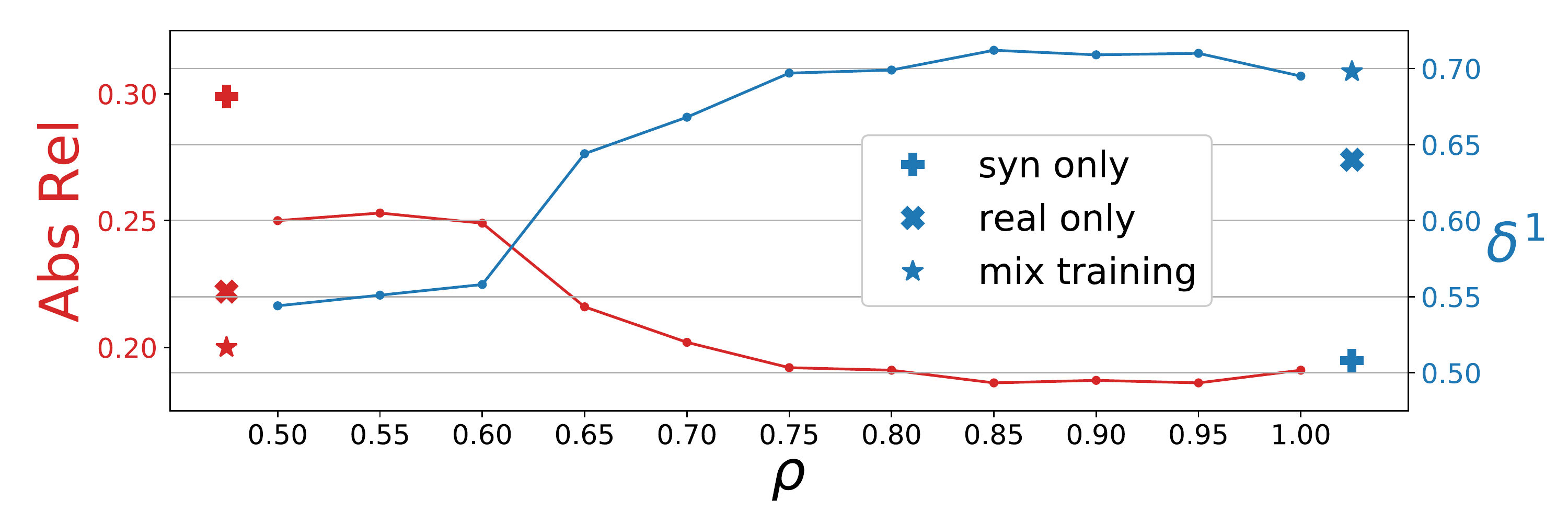}
\vspace{-7mm}
\caption{Ablation study of the sparsity factor $\rho$ of the ARC model on NYUv2 dataset.
We use ``Abs-Rel'' and ``or $\delta^1$''
to measure the performance (see Table~\ref{tab:metrics} for their definition). Note that the sparsity level $\rho$ cannot be exact 1.0 due to KL loss during training, so we present an approximate value with $\rho = 0.99999$.}
\label{fig: rho curve nyu}
\end{figure}

{
\begin{table*}[t]
\vspace{-3mm}
\centering
\tiny
\caption{\small  \textbf{Quantitative comparison} between ARC and mix training baseline inside and outside of the mask region on NYUv2 testing set~\cite{silberman2012indoor}, where $\Delta$ represents the performance gain of ARC over mix training baseline under each metric.}
\vspace{-3mm}
\resizebox{0.98\linewidth}{!}{
\begin{tabular}{l|c c c c | c c c}
\hline
\multirow{2}{*}{Model/metric}
&  \multicolumn{4}{c|}{\cellcolor{col1}\texttt{$\downarrow$ lower is better}} 
&  \multicolumn{3}{c}{\cellcolor[rgb]{0.0,0.8,1.0}\texttt{$\uparrow$ better}} \\
& \cellcolor{col1} { Rel} & \cellcolor{col1} { Sq-Rel} 
& \cellcolor{col1} { RMS} & \cellcolor{col1} { RMS-log}
& \cellcolor[rgb]{0.0,0.8,1.0} { $\delta^1$} 
& \cellcolor[rgb]{0.0,0.8,1.0} { $\delta^2$} 
& \cellcolor[rgb]{0.0,0.8,1.0} { $\delta^3$} \\
\hline
\multicolumn{8}{c}{\cellcolor{lightgrey}
Inside the mask (e.g., removed/inpainted)} \\
\hline
mix training & 0.221 & 0.259 & 0.870 & 0.282 & 0.661 & 0.889 & 0.966 \\
ARC: \emph{full} & {0.206} & {0.232} & {0.851} &
{0.273} & {0.675} & {0.895} & {0.970} \\
{\qquad $\Delta$} & $\downarrow$0.015 & $\downarrow$0.027 & $\downarrow$0.019 & $\downarrow$0.009 & $\uparrow$0.014 & $\uparrow$0.006 & $\uparrow$0.004 \\
\hline
\multicolumn{8}{c}{\cellcolor{lightgrey}
Outside the mask} \\
\hline
mix training & 0.198 & 0.191 & 0.715 & 0.256 & 0.700 & 0.913 & 0.976 \\
ARC: \emph{full} & {0.185} & {0.173} & {0.703} &
{0.249} & {0.713} & {0.918} & {0.977} \\
{\qquad $\Delta$} & $\downarrow$0.013 & $\downarrow$0.018 & $\downarrow$0.012 & $\downarrow$0.007 & $\uparrow$0.013 & $\uparrow$0.005 & $\uparrow$0.001 \\
\hline
\end{tabular}
}
\label{tab:inside outside}
\vspace{-1mm}
\end{table*}
}

\begin{figure}[t]
\begin{center}
\includegraphics[width=0.99\linewidth]{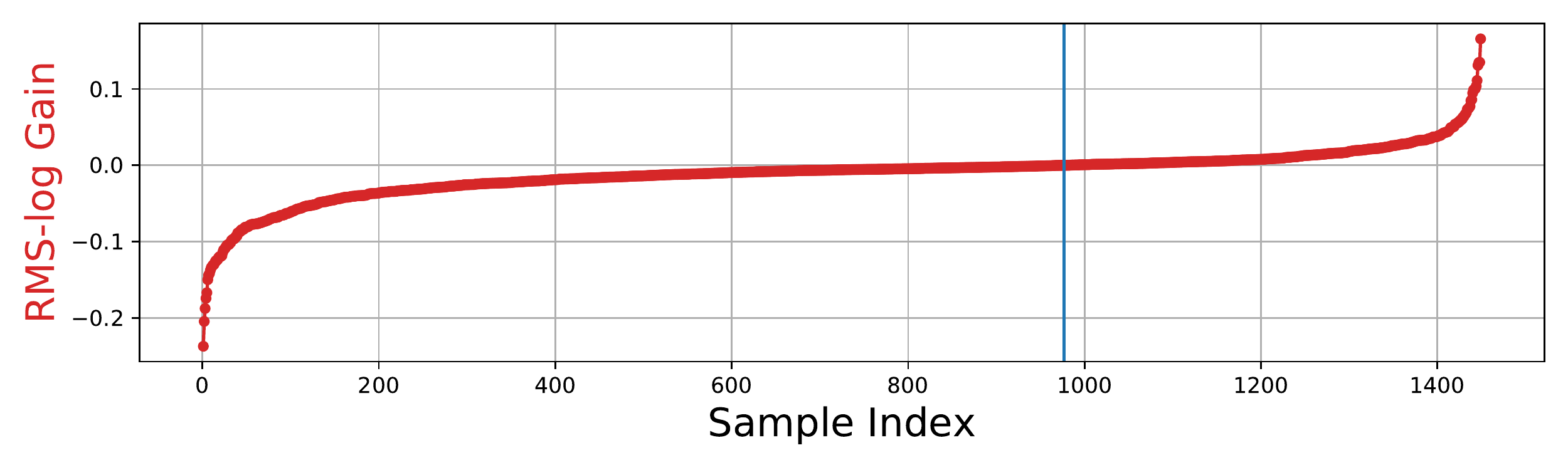}
\vspace{-3mm}
\caption{Sorted per-sample error reduction of ARC over the mix training baseline on NYUv2 dataset w.r.t the RMS-log metric. The error reduction is computed as RMS-log(ARC) $-$ RMS-log(mix training). The blue vertical line represents the index separating negative and positive error reduction.
}

\label{fig: per-sample-improvement}
\vspace{-7mm}
\end{center}
\end{figure}

\begin{figure*}[ht]
\begin{center}
    \includegraphics[width=0.995\textwidth]{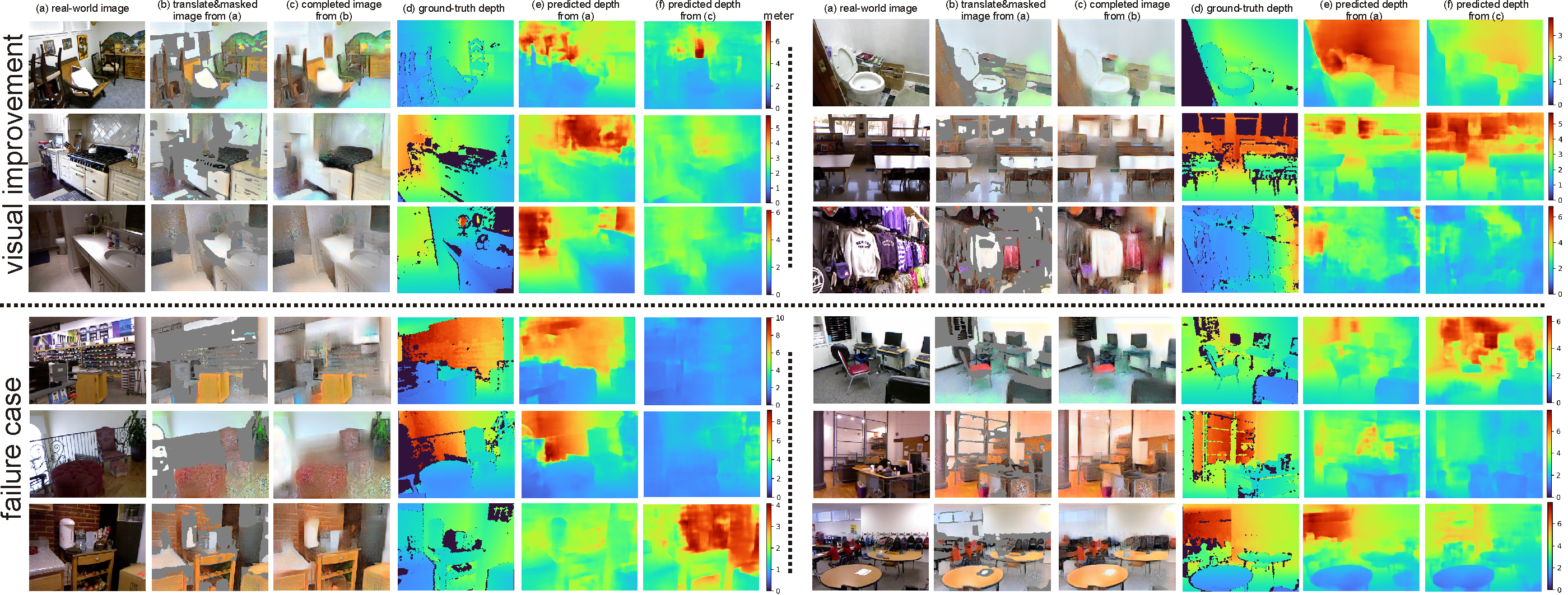}
\end{center}
\vspace{-5mm}
\caption{\small
Qualitative results list
some images over which our ARC model
improves the depth prediction remarkably,
as well as failure cases. (Best viewed in color and zoomed in.) More visualizations of ARC on Kitti and NYUv2 dataset are shown in the Appendices.
}
\vspace{-3mm}
\label{fig:visual_results}
\end{figure*}

\noindent\textbf{Sparsity level $\rho$} controls 
the percentage of pixels to remove in the binary attention map.
We are interested in studying how the hyperparameter $\rho$
affects the overall depth prediction.
We plot in Fig.~\ref{fig: rho curve nyu}  
the performance vs. varied $\rho$ on
two metrics (Abs-Rel and $\delta^1$).
We can see that the overall depth prediction 
degenerates slowly with smaller $\rho$ at first;
then drastically degrades when $\rho$ decreases to 0.8, 
meaning $\sim20\%$ pixels are removed for a given image.
We also depict our three baselines,
showing that over a wide range of $\rho$.
Note that ARC has slightly worse performance when $\rho=1.0$ (strictly speaking, $\rho=0.99999$) compared to $\rho=0.95$. This observation shows that remove a certain amount of pixels indeed helps depth predictions.
The same analyses on Kitti dataset are shown in the Appendices.

\noindent\textbf{Per-sample improvement study} to understand where our ARC model performs better over
a strong baseline. We sort and plot the per-image error reduction of ARC over the mix training baseline on NYUv2 testing set according to the RMS-log metric, shown in Fig.~\ref{fig: per-sample-improvement}. The error reduction is computed as RMS-log(ARC) $-$ RMS-log(mix training). It's easy to observe that ARC reduces the error for around 70\% of the entire dataset. More importantly, ARC decreases the error over 0.2 at most when the average error of ARC and our mix training baseline are 0.252 and 0.257, respectively. 

\noindent \textbf{Masked regions study} analyzes how ARC differs from mix training baseline on ``hard" pixels and regular pixels. For each sample in the NYUv2 testing set, we independently compute the depth prediction performance inside and outside of the attention mask. As shown in Table~\ref{tab:inside outside}, ARC improves the depth prediction not only inside but also outside of the mask regions on average. This observation suggests that ARC improves the depth prediction globally without sacrificing the performance outside of the mask.

\noindent\textbf{Qualitative results} are shown in Fig.~\ref{fig:visual_results},
including some random failure cases (measured by performance drop when using ARC).
These images are from NYUv2 dataset.
Kitti images and their results can be found in the Appendices.
From the good examples,
we can see ARC indeed removes some cluttered regions that are intuitively challenging:
replacing clutter with simplified contents, \eg,
preserving boundaries and 
replacing bright windows with wall colors. 
From an uncertainty perspective, removed pixels are places where models are less confident. By comparing with the depth estimate over the original image, ARC's learning strategy of “learn to admit what you don’t know” is superior to making confident but often catastrophically poor predictions.
It is also interesting to analyze the failure cases.
For example,
while ARC successfully removes and inpaints 
rare items like the frames and cluttered books in the shelf,
it suffers from over-smooth areas that provide little cues to infer the 
scene structure.
This suggests future research directions, 
\eg improving modules with the unsupervised real-world images,
inserting a high-level understanding of the scene with partial labels (\eg easy
or sparse annotations) for tasks in which real annotations are expensive or even
impossible (\eg, intrinsics).

\vspace{-5pt}
\section{Conclusion}
\vspace{-7pt}
We present the ARC framework which learns to attend,
remove and complete ``hard regions'' that depth predictors
find not only challenging but detrimental to overall depth prediction performance.
ARC learns to carry out completion over these removed regions in order to 
simplify them and bring them closer to the distribution of the synthetic domain.
This real-to-synthetic translation ultimately makes better use of synthetic 
data in producing an accurate depth estimate.
With our proposed modular coordinate descent training protocol,
we train our ARC system and demonstrate its effectiveness
through extensive experiments:
ARC outperforms other state-of-the-art methods in depth prediction,
with a limited amount of annotated training data
and a large amount of synthetic data.
We believe our ARC framework is also applicable of boosting performance on 
a broad range of other pixel-level prediction tasks, such as surface normals 
and intrinsic image decomposition, where per-pixel annotations are similarly
expensive to collect.
Moreover, ARC hints the benefit  of uncertainty estimation that requires special attention
to the ``hard'' regions for better prediction.

\noindent\textbf{Acknowledgements} This research was supported by NSF grants IIS-1813785, IIS-1618806, a research gift from Qualcomm, and a hardware donation from NVIDIA.

{\small
\bibliographystyle{ieee_fullname}
\bibliography{egbib}
}

\end{document}